\documentclass{article}

\ifdefined\pdfminorversion
\fi

\usepackage{arxiv}

\usepackage[utf8]{inputenc} 
\usepackage[T1]{fontenc}    
\usepackage{hyperref}       
\usepackage{url}            
\usepackage{booktabs}       
\usepackage{amsfonts}       
\usepackage{nicefrac}       
\usepackage{microtype}      
\usepackage{lipsum}
\usepackage{graphicx}
\graphicspath{ {./images/} }
\usepackage{color,array,amsthm}
\usepackage{amsmath,amsfonts}

\providecommand{\tablefont}{\small}

\title{\normalfont\bfseries nBMS, a Neuromorphic Battery Management System with a Silicon-Validated Spiking State-of-Charge Core for eVTOL Aircraft}

\author{
 \.{I}. CAN DIKMEN \\
  Department of Electrical and Electronics Engineering \\ Istinye University, Istanbul, Turkiye \\
  \texttt{can.dikmen@istinye.edu.tr} \\
}

\begin{document}

\maketitle

\begin{abstract}
State-of-charge (SoC) estimation for electric vertical take-off and landing (eVTOL) aircraft must run on the vehicle under hard energy and certification budgets, on a duty cycle unlike anything in the automotive literature. In this study an event-driven spiking network, the state-estimation core of the nBMS neuromorphic battery management architecture, is designed for per-timestep SoC estimation and evaluated on a public 22-cell eVTOL dataset with an automotive cross-check. A delta and population encoder, a second-order sigma-delta spiking layer, and a rate-accumulator readout hold 34{,}433 int16 parameters with zero dense multiply-accumulate (MAC) operations. The estimator reaches 2.45\% root-mean-square error (RMSE) against 1.74\% for a tuned long short-term memory (LSTM) baseline, a gap characterized as a temporal-mixing limit in cruise; in exchange it degrades 1.6 times slower than an adaptive leaky integrate-and-fire control under sensor noise and needs roughly 3{,}500 additions per step where the LSTM needs 67{,}700 multiply-accumulates. The full core is deployed on a low-cost automotive-qualified Artix-7 field-programmable gate array at 87\% block-RAM utilization, meets timing at 50~MHz, and reproduces the frozen fixed-point reference bit-exactly over a real 491-step flight segment on silicon. Fully annotated post-implementation analysis gives 1.30~$\mu$J per inference, an average of 0.65~$\mu$W at the 0.5~Hz mission cadence.
\end{abstract}


\section{INTRODUCTION}
T{\scshape he} estimation of battery state of charge (SoC) is one of the oldest practical problems of electrochemical storage. Coulomb counting predates lithium-ion cells, and its unbounded drift was understood just as early; model-based observers of the Kalman family brought the drift under control at the cost of per-cell parameterization, and in the last decade recurrent and convolutional networks pushed automotive accuracy below one percent in favorable conditions \cite{chemali_tie,hannan}. So, on the algorithmic side the automotive problem looks mature.

Electric vertical take-off and landing (eVTOL) aircraft break this picture twice. The duty cycle is different in kind: a high-rate hover, a calm cruise, and a second high-rate landing, at depths of discharge automotive cycles never reach; the public eVTOL campaign of Bills et al.\ \cite{bills} was collected precisely because automotive data does not transfer. And the energy and certification budget of an airborne battery unit is far tighter than a ground vehicle's, so an estimator burning a hundred thousand multiply-accumulate (MAC) operations per step is a real cost item, whereas an event-driven network adds weights only when spikes occur. The same aerial platforms are meanwhile adopting event-based sensing for perception and navigation, and an event-driven battery core keeps the management function inside that regime instead of anchoring it to a clocked island. It is seen that aviation invites exactly the model class the automotive problem never forced anyone to adopt. However, the published record on spiking battery estimation is thin, dominated by health-oriented work, and to the authors' knowledge contains no per-timestep SoC estimator on an aviation duty cycle, let alone one validated on silicon.

In this study an event-driven spiking estimator is designed, trained, and evaluated for per-timestep SoC on the eVTOL dataset, with an automotive dataset based on the Worldwide Harmonized Light Vehicles Test Procedure (WLTP) as a cross-domain check \cite{WLTPData}. The estimator constitutes the state-estimation core of nBMS, a patented neuromorphic battery management architecture integrating spike encoding, a spiking core, decision post-processing, and gate drive on one die \cite{nbms_patent}; the aviation configuration, spanning eVTOL aircraft and smaller unmanned platforms, is designated nBMS-Aero, and this work is its first silicon-validated core. The neuron is deliberately taken from the established sigma-delta family \cite{boeshertz}, and an adaptive leaky integrate-and-fire (LIF) ablation is reported in full. The contribution is the intersection: the first event-driven per-timestep SoC estimator on an aviation duty cycle, a deployable operating point with zero dense multiply-accumulates, a reframing of SoC regression as oversampled one-bit quantization that explains the observed noise robustness, and a complete silicon validation on a low-cost automotive-qualified field-programmable gate array (FPGA), bit-exact over a real flight segment.

On clean data the spiking estimator does not beat a well-tuned long short-term memory (LSTM) network, the adaptive-LIF control wins on clean accuracy while the second-order neuron wins under noise, both facts carry paired statistics, and the primary variant was pre-registered before the final evaluation with the timeline disclosed in Section~IV. For this reason we consider the negative findings part of the contribution. Section~II positions the work, Sections~III and~IV give the core and the protocol, Section~V reports the results, Section~VI presents the hardware and its validation, and Sections~VII and~VIII discuss and conclude.

\section{RELATED WORK}
\subsection{State-of-Charge Estimation}
Model-based estimation built the field; equivalent-circuit models with Kalman-family observers remain the embedded workhorse. Data-driven estimators then took the accuracy frontier: the LSTM line of Chemali et al.\ reports 0.57\% mean absolute error (MAE) at fixed temperature and 1.10\% with a plain deep network \cite{chemali_tie}, self-supervised Transformers reach 0.90\% root-mean-square error (RMSE) \cite{hannan}, temporal convolutions follow as the second canonical family \cite{herle}, and residual convolutional neural network (CNN) variants reach 1.26\% MAE while reporting $2.24\times10^{6}$ floating-point operations per estimate \cite{rescnn}. That last number is telling: the accuracy race is largely settled and competition has moved to deployment cost, with pruned lightweight SoC networks in the transportation-electrification literature \cite{pruned_tte} and system-level validation inside resource-constrained battery management system (BMS) units \cite{embedded_bms}. All of these remain clock-driven dense networks. On the aviation side the model-based line stays active where data is thin: in this journal, Uhm and Kim couple an adaptive robust extended Kalman filter to remaining-flying-time prediction for a small unmanned aerial vehicle \cite{uhm_taes}, Goshtasbi et al.\ trace how SoC initialization error propagates into the state-of-power limits of an eVTOL pack \cite{goshtasbi}, and the review of Raoofi and Yildiz maps the estimation and airworthiness questions for aircraft propulsion batteries \cite{raoofi}. On the data side the Bills campaign \cite{bills} is the anchor dataset; modeling on it so far targets health and life with conventional machine learning \cite{granado}, and the one SoC study we are aware of applies tree and kernel models on a single random split \cite{evtol_ml_soc}. So per-timestep SoC under a leave-cells-out protocol on an aviation duty cycle is, comparing to the automotive case, essentially unexplored.

\subsection{Spiking Networks for Continuous-Valued Time Series}
Spiking networks are overwhelmingly a classification technology; continuous regression from binary spikes is a recognized hard problem with few precedents. Henkes et al.\ demonstrated nonlinear regression with membrane-potential decoding \cite{henkes} and SeqSNN adapted forecasting architectures to the spiking domain at ICML scale \cite{seqsnn}; the common ingredient is a non-spiking leaky readout in the LSNN tradition \cite{bellec}, also adopted here. On encoding, the survey of Auge et al.\ \cite{auge} covers the standard schemes and a recent comparison found multi-threshold delta modulation the most noise-robust \cite{enc_bench}; this work combines exactly that signed scheme with a Gaussian population code. Neuron models beyond the plain LIF are an active line, from adaptive thresholds in the adaptive-LIF (ALIF) family \cite{bellec} to adaptive-reset and related generalized-LIF variants \cite{arlif}. The second-order neuron used here belongs to the sigma-delta cascade-of-integrators family of Boeshertz et al.\ \cite{boeshertz}, whose state-space reading was recently made explicit \cite{karilanova}; the ALIF control is the most proper in-family baseline.

\subsection{Neuromorphic Battery Management and Hardware Energy}
The intersection of spiking networks and battery estimation is thin but real. SpikeSOH estimates state of health (SoH) on nineteen lithium iron phosphate (LFP) cells, reporting 4.5\% MAE and 0.36~mJ as a proxy estimate rather than a measurement \cite{spikesoh}; SSA-Net brings spiking attention to SoH from impedance spectra \cite{ssanet}; a residual spiking detector addresses pack faults \cite{snn_fault}. It is observed that all target health or fault quantities; to the authors' knowledge no peer-reviewed spiking per-timestep SoC estimator exists, aside from a reservoir-computing proof of concept in abstract form, and none of the cluster deploys on hardware. Their energy claims rest on synaptic-operation counts priced with the Horowitz figures \cite{horowitz}, a methodology criticized directly: Lemaire et al.\ show spike-count proxies can even reverse the efficiency ranking once memory traffic is counted \cite{lemaire}. FPGA acceleration of spiking networks is meanwhile well developed, from Spiker+ on the same Artix-7 family \cite{spikerplus} to recent energy-focused recurrent cells \cite{harmeling}, but never for battery state. The present work sits in that unoccupied intersection, with the energy figure taken from fully annotated post-implementation analysis instead of an operation-count proxy.

\begin{figure*}[ht]
\centering
\includegraphics[width=\textwidth]{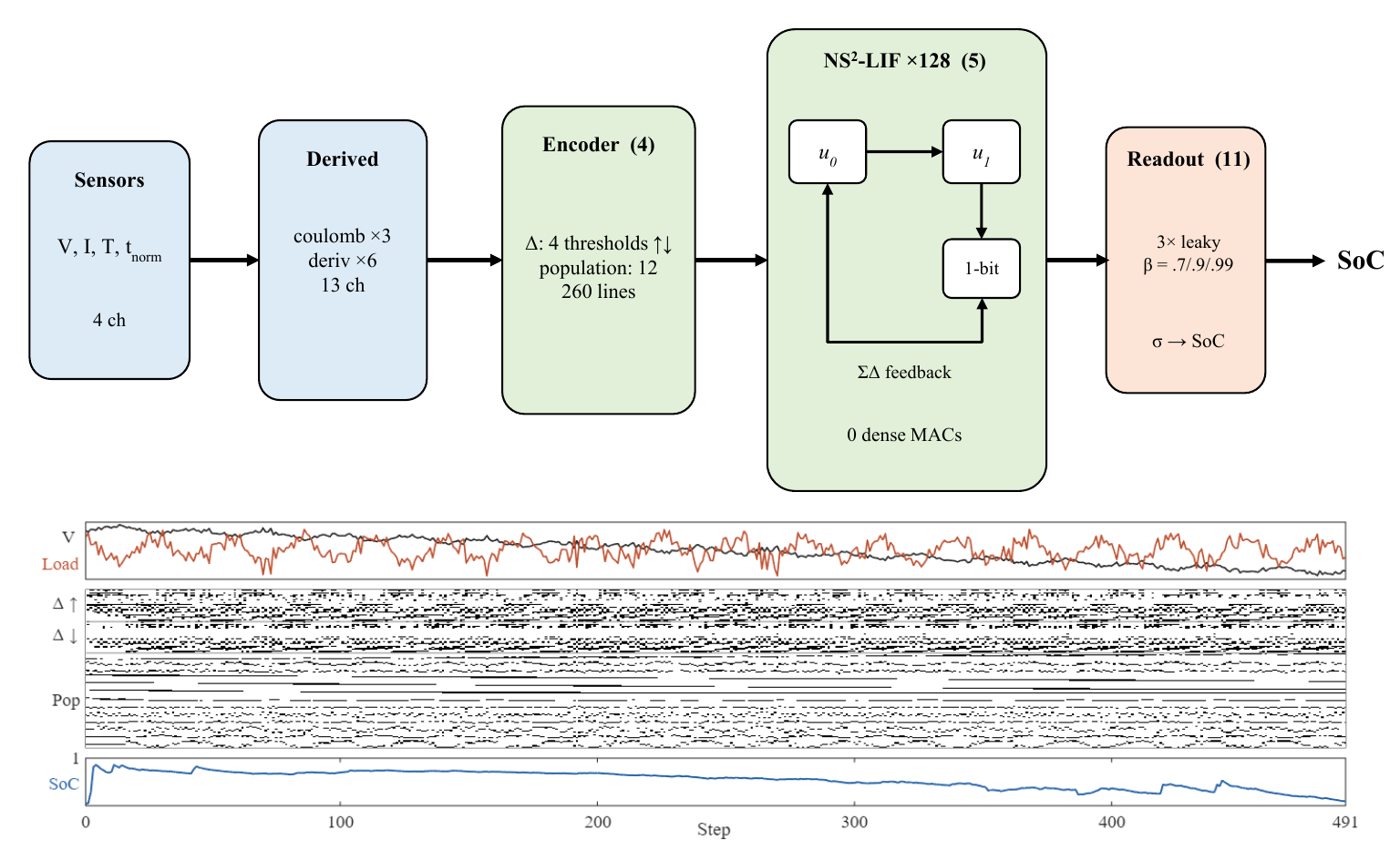}
\caption{The nBMS state-of-charge core (top; parenthesized numerals follow the architecture blocks of \cite{nbms_patent}), together with its recorded activity over the silicon-validated 491-step flight segment (bottom): the input traces, the 260-line encoder spike raster produced by the deployed fixed-point arithmetic, and the per-timestep SoC output. All three strip panels show real data from the validation vector, not schematic illustration.}
\label{fig:pipeline}
\end{figure*}

\section{THE nBMS STATE-OF-CHARGE CORE}
The estimator, shown in Fig.~\ref{fig:pipeline}, maps the four raw sensor channels of one timestep, cell voltage, current, temperature, and a normalized mission-time auxiliary, to a SoC value for the same timestep: thirteen derived channels, a hybrid encoder producing 260 binary spike lines, one hidden layer of 128 second-order spiking neurons, and a multi-timescale leaky readout ending in a sigmoid. In the terminology of the nBMS architecture \cite{nbms_patent}, the encoder of Section~III-A realizes the spike-encoder block and the spiking layer with its readout realizes the network core and the post-decision unit. The whole model holds 34{,}433 parameters: 33{,}280 input synapses with 128 biases, 385 readout weights, and 640 per-neuron dynamics parameters. Every learned operation is either a synapse applied to binary spikes or a leaky integrator, so the count of dense multiply-accumulate operations per step is exactly zero; this claim is made precise in Section~III-C.

\subsection{Input Channels and Spike Encoding}
All channels are min-max normalized to $[0,1]$ with training-cell statistics. The feedforward layer cannot integrate over long horizons by itself, so the charge history is supplied explicitly. With the per-step load defined as $\ell_t = 1 - I_t$, three charge channels are computed as
\begin{equation}
c^{\mathrm{full}}_t = \mathrm{clip}\Big(\tfrac{1}{204.8}\sum_{\tau \le t} \ell_\tau,\, 0,\, 1\Big),\quad
c^{\beta}_t = (1-\beta)\, e^{\beta}_t,
\end{equation}
where $e^{\beta}_t = \beta e^{\beta}_{t-1} + \ell_t$ and $\beta \in \{0.99, 0.90\}$. The full integral carries the slow cruise drift, the leaky pair act as recent-charge and polarization proxies. Additionally, six derivative channels make the takeoff and landing voltage-sag shape explicit: for horizons $h \in \{1,4,16\}$ and both signals $z \in \{V, \ell\}$,
\begin{equation}
d^{z,h}_t = \tfrac{1}{2} + \tfrac{1}{2}\tanh\big(4\,(z_t - z_{t-h})\big),
\end{equation}
centered at $0.5$ when the signal is flat. The encoder then combines change and level coding uniformly over the thirteen channels: a signed multi-threshold step-forward code with thresholds $\theta \in \{0.01, 0.03, 0.08, 0.20\}$ maintains one reference value per threshold, initialized at the first sample, and emits
\begin{equation}
s^{\uparrow}_t = \mathbf{1}[x_t - r_{t-1} \ge \theta],\qquad
s^{\downarrow}_t = \mathbf{1}[r_{t-1} - x_t \ge \theta],
\end{equation}
with reference tracking $r_t = r_{t-1} + \theta(s^{\uparrow}_t - s^{\downarrow}_t)$. Change coding is blind to absolute level, so a Gaussian population code over twelve uniform centers $c_k$ is added, binarized at the half-maximum of the receptive field:
\begin{equation}
p^{k}_t = \mathbf{1}\big[\,|x_t - c_k| \le \sigma\sqrt{2\ln 2}\,\big],\qquad \sigma = 0.06 .
\end{equation}
Each channel contributes $4\times2 + 12 = 20$ lines, 260 in total. It can be seen that the event rate follows the duty cycle by construction: the delta lines are nearly silent in cruise and burst at takeoff and landing, while the population lines fire only near the current level.

\subsection{Second-Order NS\texorpdfstring{\textsuperscript{2}}{2}-LIF Neuron}
The hidden layer uses a two-state cascade-of-integrators neuron with output-spike feedback into both states. With $I_t = W s_t + b$ the synaptic drive from the encoder spikes, each neuron updates as follows,
\begin{align}
u^{(0)}_t &= \mathrm{clip}\big(a_0\, u^{(0)}_{t-1} + I_t - k_0\,\vartheta\, s_{t-1},\ \pm 8\big),\\
u^{(1)}_t &= \mathrm{clip}\big(a_1\, u^{(1)}_{t-1} + g\, u^{(0)}_t - k_1\,\vartheta\, s_{t-1},\ \pm 8\big),\\
s_t &= \mathbf{1}\big[u^{(1)}_t \ge \vartheta\big],\qquad \vartheta = 1,
\end{align}
where $a_0, a_1 \in (0,1)$ are per-neuron leak factors parameterized through a sigmoid, and $g, k_0, k_1 > 0$ are per-neuron coupling and feedback gains parameterized through a softplus. The binary output feeds back into both stages, so the loop is a second-order sigma-delta modulator on the synaptic drive; the NS\textsuperscript{2} name stands for this second-order noise shaping. The neuron is an instantiation of the established family of \cite{boeshertz}, with the state-space reading of \cite{karilanova}. Spiking is deterministic; training uses backpropagation through time with a fast-sigmoid surrogate of slope 25. In order to keep the comparison fair, the same pipeline with an adaptive-threshold LIF \cite{bellec} in place of this neuron is trained identically and reported in full as the internal control.

\subsection{Rate-Accumulator Readout}
The readout decodes a continuous value from the hidden spike vector $s_t$ without any dense layer on continuous activations. Three leaky accumulators with time constants spanning roughly three, ten, and a hundred steps integrate synaptic projections of the spikes,
\begin{equation}
r^{(i)}_t = \beta_i\, r^{(i)}_{t-1} + W^{ro}_i\, s_t,\qquad \beta_i \in \{0.70, 0.90, 0.99\},
\end{equation}
and the estimate is $\hat{y}_t = \mathrm{sigmoid}\big(\sum_i r^{(i)}_t + b^{ro}\big)$. Here, a linear map of a leaky average equals the leaky average of the mapped spikes, so the readout is exactly accumulate-on-spike followed by scalar leaks, and the zero-dense-MAC property is an identity, not an approximation. Output is per timestep, with no decoding window.

\subsection{Fixed-Point Specification}
The deployed arithmetic is frozen in Table~\ref{tab:goldspec}, locked before any hardware work and validated by a pure-integer golden model mirroring the register-transfer dataflow operation for operation; the leak factors $a$ stay in Q1.15 while $g$ and $k$ share the membrane format, and the charge integral is exact since $204.8$ admits the form $\times 160 \gg 15$. An int8 weight format was evaluated first and rejected: per-tensor int8 raised validation RMSE by 3.45 SoC points, and per-row scaling, while recovering most of the accuracy, still flipped 3.4\% of the hidden spikes against the float reference, which a bit-exact deployment contract cannot accept. With int16 the integer model tracks the float reference within 0.07 points, at the lookup-table (LUT) resolution floor, and the 67~kB weight memory makes block RAM (BRAM) the binding constraint of Section~VI.

\begin{table}[!t]
\caption{Frozen Fixed-Point Specification (GOLD-1)}
\label{tab:goldspec}
\centering
\tablefont
\begin{tabular}{ll}
\toprule
Weights $W_{in}$, $W_{ro}$ & int16, Q1.15 \\
Membrane states & Q13.18 (32-bit), clamp $\pm8$ \\
LUTs (tanh / sigmoid) & $8193{\times}16$\,b / $2049{\times}16$\,b \\
Rounding; coulomb & nearest; $\times160{\gg}15$ (exact $/204.8$) \\
Parameters / dense MACs & 34{,}433 int16 / 0 per step \\
\bottomrule
\end{tabular}
\end{table}

\section{EXPERIMENTAL SETUP}
\subsection{Datasets and Evaluation Protocols}
The primary dataset is the eVTOL campaign of Bills et al. \cite{bills}: twenty-two Sony-Murata VTC-6 cells under repeated takeoff, cruise, and landing missions. Per cell at most 300 cycles are used, subsampled uniformly over the cell's life, and each mission is stride-downsampled to at most 512 steps keeping the whole profile intact. The target is per-timestep SoC; all figures are RMSE in SoC percentage points; cruise and transient regimes split at 0.35 on the normalized load.

Two protocols are used: a fixed primary split holding five of the twenty-two cells out entirely (one fixed-seed shuffle; the five are never seen), and a five-fold leave-cells-out (LOCO) cross-validation with round-robin folds of four to five cells, every model retrained per fold. The fixed split carries the headline comparison; the cross-validation carries the generalization claim. To the authors' knowledge the Bills campaign is the only public battery dataset collected under an aviation mission profile with laboratory-grade ground truth, so the out-of-domain check is automotive by necessity: a secondary WLTP-based dataset of twelve cells serves strictly as a cross-domain check, twelve-fold leave-one-cell-out, with no design decision taken on it \cite{WLTPData}.

Input-noise robustness is evaluated by adding zero-mean Gaussian noise to all normalized input channels of the held-out cells at $\sigma \in \{0, 0.01, 0.02, 0.05, 0.1\}$ and recording the ensemble RMSE at each level; the degradation factor reported in Section~V is the ratio of the noisy to the clean RMSE.

\subsection{Models, Training, and Baselines}
All spiking variants share the Section~III pipeline and one recipe: AdamW ($10^{-3}$, weight decay $10^{-5}$), batch 256, cosine decay over a 120-epoch horizon capped at 140, patience-25 early stopping, full backpropagation through time over 512 steps, masked mean-squared error (MSE) on per-timestep SoC. A deployable noise augmentation was locked before the final runs: Gaussian 0.002 and 0.010 on normalized voltage and current plus an impulsive 0.2-amplitude spike at probability 0.01 as an electromagnetic-interference (EMI) proxy; it costs nothing at deployment and essentially nothing on clean accuracy.

Two external baselines train on identical data: a single-layer LSTM of the Chemali line \cite{chemali_tie} with width 128, and a small causal temporal convolutional network (TCN) with kernel 3 and four levels, both tuned with a small grid over learning rate and weight decay, six configurations each with the winner confirmed over five seeds, and retrained from scratch with the final seeds.
The internal control is the same pipeline with the neuron swapped for an ALIF \cite{bellec}, trained identically including the augmentation. The control exists because of a rule this project adopted after an earlier failure: any nonstandard component must face its trivial swap at full training before further effort goes into it; external baselines do not answer the component-level question.

Every reported model is a five-seed ensemble (fixed seeds, independent training, mean prediction deployed); single-seed numbers feed the paired statistics. Wilcoxon signed-rank tests over paired folds give the significance figures, with the caveat that the five-fold eVTOL protocol has limited power and the twelve-fold automotive protocol is the sharper instrument.

\subsection{Pre-Registration and Selection Disclosure}
The timeline matters for reading Section~V. The NS\textsuperscript{2}-LIF variant was designated primary and frozen before the cross-domain and noise evaluations were finalized, with the ALIF control specified at the same time and the commitment to report it in full whichever way it fell. It fell against us on clean data, and those numbers are in every table of this paper.

A second disclosure concerns tuning. A 25-trial TPE search over the NS\textsuperscript{2} configuration improved the primary split but worsened leave-cells-out RMSE from 2.818 to 3.381, which we read as overfitting the selection objective to the five validation cells; it was rejected. So the primary variant is reported untuned while both baselines are reported tuned. This asymmetry favors the baselines and is intentional.

\section{RESULTS}
\begin{table}[ht]
\caption{Accuracy, Robustness, and Cost (\% SoC RMSE; 5-Seed Ensembles)}
\label{tab:results}
\centering
\tablefont
\setlength{\tabcolsep}{4pt}
\begin{tabular}{@{}lcccc@{}}
\toprule
 & Prim. & LOCO (5f) & Deg.\,@\,.05 & Ops/step \\
\midrule
NS\textsuperscript{2} & 2.449 & $2.82\pm0.97$ & $\times3.81$ & 3.5k add \\
ALIF & 2.073 & $3.26\pm1.40$ & $\times6.23$ & 3.5k add \\
LSTM & 1.736 & 3.203 & $\times1.62$ & 67.7k MAC \\
TCN & 1.707 & 2.640 & $\times1.93$ & 149k MAC \\
\midrule
\multicolumn{5}{@{}p{0.97\columnwidth}@{}}{\footnotesize LSTM/TCN are TPE-tuned; the spiking variants are untuned (Sec.~IV-C). Degradation is on KIT. Paired eVTOL LOCO, NS\textsuperscript{2} vs ALIF: $p{=}0.312$; KIT LOCO (12 folds): ALIF 11/12, $p{=}0.0015$.} \\
\bottomrule
\end{tabular}
\end{table}

Table~\ref{tab:results} carries the clean-accuracy ordering, and it is reported control-first: on the primary split the ALIF control reaches 2.073\% against 2.449\% for the pre-registered NS\textsuperscript{2}-LIF, while the tuned LSTM and TCN lead at 1.736\% and 1.707\%. The gap to the baselines is concentrated in cruise, where the spiking estimator stays near 2.0\% and the LSTM reaches roughly 1.4\%; the transient segments are close. So the clean-data ranking is dense baselines first, ALIF second, NS\textsuperscript{2} third, and none of the later sections revises it.

Cross-validation adds two qualifications. On the five-fold eVTOL protocol the ensembles give $2.818\pm0.97$ for NS\textsuperscript{2}, $3.263\pm1.40$ for ALIF, 3.203 for the LSTM, and 2.640 for the TCN; the spiking ordering nominally reverses, with NS\textsuperscript{2} ahead in four of five folds, but the paired difference is not significant ($p=0.312$) and is driven almost entirely by a single fold, outside which the means differ by 0.05. It is the twelve-fold automotive protocol that resolves the pair: ALIF wins eleven of twelve folds ($p=0.0015$) at $2.470\pm0.25$ against $2.624\pm0.36$, and the TCN wins all twelve against NS\textsuperscript{2} ($p=0.0005$). The clean-accuracy verdict of Table~\ref{tab:results} therefore stands on the primary split and on the sharper cross-domain instrument, not on the noisy eVTOL folds.

\begin{figure}[ht]
\centering
\includegraphics[width=0.7\textwidth]{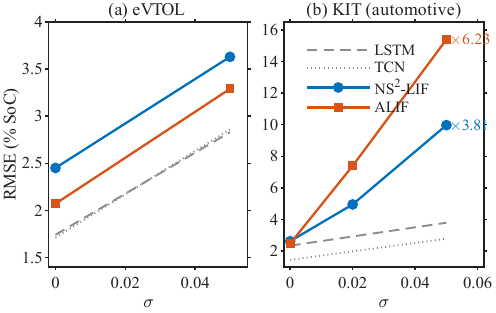}
\caption{RMSE under Gaussian input noise on all normalized channels: (a) eVTOL, (b) KIT automotive.}
\label{fig:noise}
\end{figure}

\begin{figure}[ht]
\centering
\includegraphics[width=0.7\textwidth]{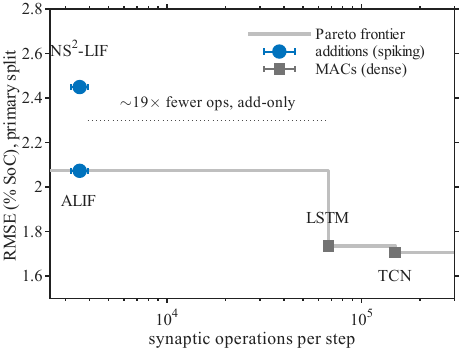}
\caption{Primary-split accuracy against synaptic operations per step (5-seed ensembles); the gray staircase is the Pareto frontier.}
\label{fig:pareto}
\end{figure}

Under input noise the picture inverts, and Fig.~\ref{fig:noise} shows where. On the eVTOL panel the two spiking variants separate only mildly ($\times1.48$ against $\times1.59$ at $\sigma=0.05$). On the automotive panel the separation is wide: the ALIF error climbs through 7.39 at $\sigma=0.02$ to 15.39 at $\sigma=0.05$, a factor of 6.23 over its clean value, while the second-order neuron reaches 9.97, a factor of 3.81, a 1.6-times slower breakdown consistent with the sigma-delta reading of Section~III-B. The dense baselines, which consume continuous inputs and carry no spike encoder, degrade least of all; the robustness claim of this paper is strictly the in-family comparison. Efficiency completes the trade. Fig.~\ref{fig:pareto} places the four models on the accuracy-cost plane: the spiking variants operate at 3.2 to 3.9 thousand additions per step where the LSTM spends 67{,}700 multiply-accumulates and the TCN 149{,}000, so the 0.34 to 0.71 point accuracy premium of the dense models is bought at roughly nineteen to forty times the operation count, in multiplications rather than additions. The Pareto frontier of the figure runs through ALIF, the LSTM, and the TCN; NS\textsuperscript{2} sits off the clean-data frontier and joins it only when the noise axis of Fig.~\ref{fig:noise} is priced in.

\section{FPGA IMPLEMENTATION AND SILICON VALIDATION}
The core was implemented in Verilog and deployed on a Digilent Arty A7-35T carrying the automotive-qualified xc7a35ticsg324-1L. In order to keep the silicon bit-exact against the training side, a frozen golden artifact was generated first, quantized weights, both activation tables, and a 491-step input vector from a real flight segment with its expected per-step outputs, and every later verification layer was measured against it.

One estimation step is a linear pass of a thirteen-state machine, preprocessing, encoding, an event-driven accumulation that walks only the active spike lines, a two-phase neuron update over the 128 units, and a readout ending in a sigmoid lookup, in 1{,}227 clock cycles. There is no dense matrix engine; the four digital signal processing (DSP) blocks of Table~\ref{tab:hw} implement the scalar readout and neuron products, and synaptic integration is pure spike-gated addition. Eight parallel block-RAM banks of weights dominate the budget: the core fits the smallest Artix-7 at 87\% BRAM and 57\% LUT, so the binding constraint is memory, not logic. The board build closes timing at 50~MHz with $+0.058$~ns slack; the unpipelined second-order chain, two cascaded DSP multipliers into a 32-bit accumulate and clamp, bounds Fmax near 57~MHz, and with 24.5~$\mu$s per step against a two-second cadence, pipelining it was deliberately not pursued.

Power was estimated post-implementation with full switching-activity annotation: a gate-level functional simulation replayed 64 golden steps at the deployed 50~MHz clock, the switching activity interchange format (SAIF) annotation covered 100\% of the 33{,}442 nets at high confidence, and the 53~mW dynamic power over the 24.54~$\mu$s step gives 1.30~$\mu$J per inference. Here, the window matters: an 8-step window from the segment start gives 1.06~$\mu$J because the long-horizon derivative channels have not yet activated, so the 64-step steady state is reported. For scale, the closest published spiking battery-estimation energy figure is the 0.36~mJ per inference reported for SpikeSOH on an embedded processor \cite{spikesoh}; the task and the platform differ, but the present core sits more than 270 times lower, and its figure comes from the routed netlist rather than from an operation-count price. At the 0.5~Hz cadence the average dynamic draw is 0.65~$\mu$W, five orders below the 61~mW static of this SRAM FPGA family. It is concluded that the estimator is ready for the microwatt class while the platform is not; a flash FPGA or application-specific integrated circuit (ASIC) is the carrier, the subject of a follow-up with rail-level measurement.

\begin{table}[ht]
\caption{Implementation Results, xc7a35ticsg324-1L (OOC: Out of Context; WNS: Worst Negative Slack)}
\label{tab:hw}
\centering
\tablefont
\begin{tabular}{lcc}
\toprule
 & Core (OOC) & Board (top) \\
\midrule
LUT & 11{,}804 (56.8\%) & 10{,}046 (48.3\%) \\
Flip-flops & 12{,}503 (30.1\%) & 12{,}333 \\
Block RAM tiles & 43.5 (87\%) & 43.5 (87\%) \\
DSP48 & 4 & 4 \\
Clock / WNS & -- & 50\,MHz / $+0.058$\,ns \\
\midrule
Cycles per step & \multicolumn{2}{c}{1{,}227 (24.54\,$\mu$s)} \\
$P_{dyn}$ / $P_{static}$ & \multicolumn{2}{c}{53\,mW / 61\,mW} \\
Energy per step & \multicolumn{2}{c}{1.30\,$\mu$J (dynamic)} \\
\bottomrule
\end{tabular}
\end{table}

\subsection{Verification Methodology and a Silicon-Only Failure}
Three simulation layers preceded the board, behavioral register-transfer-level (RTL) simulation over all 491 steps, post-implementation functional, and post-implementation timing, all bit-exact against the golden reference, and all three passed. On silicon the first bitstream diverged deterministically; a rebuild with a different top diverged deterministically at exactly step 187, robust to input pacing; a third, instrumented build diverged from step 0. Three builds, three distinct deterministic failures while every simulation passed pointed at a synthesis-sensitive construct rather than at the logic.

For this purpose a state-readback instrument was added, a second universal asynchronous receiver-transmitter (UART) frame returning any internal state word including the live weight RAMs, so the complete architectural state after any step could be compared word-by-word against the golden model. The comparison localized the fault at once: preprocessing, encoder references, and coulomb accumulators matched exactly, while the synaptic accumulators of specific neurons carried values near $10^{8}$, two orders beyond what correct accumulation can produce. The failing construct was eight read-modify-write operations with different dynamic indices into one register array in a single clock; synthesis built this multi-port structure differently, sometimes incorrectly, in every context, explaining the three signatures. Splitting the array into eight per-bank files with one read-modify-write port each removed the fault, and the final bitstream reproduced the golden reference over the full 491-step segment on the board, every output word exact.

Two observations carry over. Bit-exact hardware-in-the-loop comparison against a frozen reference caught a fault class three layers of gate-level simulation could not, since the fault existed only in specific synthesis outcomes; and multi-ported read-modify-write into inferred register arrays should be treated as a hazard pattern in neuromorphic accumulator design, where per-bank splitting is cheap and removes the ambiguity.

\section{DISCUSSION}
The two spiking variants split the evaluation along two axes, and stating the split plainly is more useful than declaring a winner. On clean data the adaptive-threshold control is better: eleven of twelve automotive folds at $p=0.0015$, and the lead on the primary split. Under injected noise the ordering reverses widely: at $\sigma = 0.05$ the second-order neuron degrades by 3.81 against 6.23, a 1.6-times slower breakdown, exactly what the sigma-delta reading of Section~III-B predicts, since the output-spike feedback closes a second-order noise-shaping loop around the one-bit quantizer and pushes perturbations away from the slow band the SoC signal occupies; the adaptive-threshold neuron has no comparable loop. The deployment guidance follows from the sensor budget: filtered, well-conditioned measurements favor the control, raw shunt and thermistor lines in an electrically noisy airframe favor the second-order neuron. Both fit the same hardware.

The cruise segment deserves its own paragraph because it is where the limit of this model class shows. No configuration we probed brings the spiking estimator below roughly 2.0\% RMSE in cruise, while the LSTM reaches about 1.4\% there through continuous temporal integration. It is seen that the gap is not a tuning artifact: the TPE search of Section~IV-C improved the selection objective and worsened cross-validation from 2.818 to 3.381, so the limitation sits deeper. Our reading is a temporal-mixing gap. In cruise the encoder is nearly silent by design, the information arrives as a slow drift, and a binary spike code at this scale has limited capacity to mix information across long horizons; the explicit charge channels recover part of it, and the rest is the price of the representation. We state this as a property, not as a defect to be excused, and closing part of it with hybrid temporal features that keep the accumulate-on-spike synapse is in our view the most concrete follow-up this work points to.

Against the baselines the positioning is a Pareto argument, not an accuracy argument. The LSTM is 0.7 SoC points better and costs 67{,}700 multiply-accumulates per step, the TCN marginally better still at 149{,}000; the spiking estimator answers with 3{,}200 to 3{,}900 additions, no multiplier in the synaptic path, int16 throughout, and a bit-exact deterministic core a certification process can reason about. At 0.5~Hz none of the four is compute-bound, so the relevant budget is energy and verification, and there the operating point of Section~VI is, to our knowledge, unoccupied.

Four limitations bound the claims. First, the eVTOL cross-validation is noisy: the paired difference between the two spiking variants is not significant over the five folds ($p=0.312$) and is driven largely by a single fold, so the clean-accuracy ordering on this dataset rests on the primary split and on the sharper twelve-fold automotive protocol. Second, there is one dataset per domain, and the aviation data is bench cycling under a mission profile, not flight telemetry; transfer to an instrumented airframe is untested, and the diversity of aviation mission profiles cannot be probed with the public data available today. Third, the power figure is a post-implementation estimate. The switching activity is fully annotated and the vendor tool reports high confidence, but no current was measured on a rail, and the microwatt average at mission cadence is conditional on a platform whose static draw does not dominate, which the SRAM FPGA used here does not satisfy. A rail-instrumented measurement on a Zynq UltraScale+ platform is the subject of the follow-up study. Fourth, the silicon validation covers one full real flight segment of 491 steps; it is a strong equivalence check of the arithmetic, not a statistical statement about coverage.

\section{CONCLUSION}
Electric aviation needs estimators that respect the energy and certification budget of the airframe, and event-driven networks are a candidate the automotive literature never had to take seriously. In this study a spiking estimator with 34{,}433 int16 parameters and zero dense multiply-accumulates was evaluated on a public eVTOL dataset with an automotive cross-check and deployed on a low-cost automotive-qualified FPGA. It reaches 2.45\% RMSE against 1.74\% for an LSTM, a temporal-mixing gap reported as such; under sensor noise the second-order neuron degrades 1.6 times slower than the adaptive-LIF control, consistent with its sigma-delta coding; and the silicon reproduced the frozen fixed-point reference bit-exactly over a 491-step real flight segment at 50~MHz, at 1.30~$\mu$J per inference from fully annotated post-implementation analysis. So we consider deployability demonstrated end to end, from training checkpoint to silicon, rather than argued from proxy metrics. The sub-microwatt average draw at mission cadence is exactly the operating regime that the event-gated sleep and wake mechanism of the nBMS architecture \cite{nbms_patent} is designed to exploit, and integrating the validated core behind that mechanism is the natural next step of the nBMS-Aero line. On the hardware side, future work targets rail-instrumented power measurement on a Zynq UltraScale+ platform, where the estimate of Section~VI becomes a measured number, and a flash-FPGA port on which the static floor of the present device no longer masks the microwatt regime. On the estimation side, the same encoder and core retarget naturally to state of health, since the eVTOL dataset already carries the capacity-fade labels; we intend to close part of the cruise gap with hybrid temporal features that keep the accumulate-on-spike synapse; and the event-rate signature of the encoder is itself a candidate fault indicator in the sense of the protective functions of the architecture. A reduced configuration of the core, which our pruning probes suggest is nearly free at reduced width, is the entry point of the implant and wearable class of the same family. The balancing and gate-drive functions of the architecture remain outside the scope of this line and belong to the system-level work.

\bibliographystyle{unsrt}
\bibliography{References}

\end{document}